\documentclass[journal]{IEEEtran}
\usepackage[T1]{fontenc}
\usepackage[utf8]{inputenc}

\usepackage{amsmath,amssymb,amsthm,mathtools}
\usepackage{graphicx,booktabs,array,tabularx}
\usepackage{xcolor}
\usepackage{cite}
\usepackage[hyphens]{url}
\usepackage[hidelinks]{hyperref}
\usepackage{orcidlink}

\theoremstyle{plain}

\theoremstyle{definition}

\title{Pruned or Rewritten? Tracing Declarations Across Three Versions of a 17-Million-Line Lean Proof of the Riemann Hypothesis, Based on Eduinformatics}

\author{\mbox{Kunihiko~Takamatsu~\orcidlink{0000-0003-4172-7612}}, \mbox{Katsuhiko~Murakami~\orcidlink{0000-0002-1923-4259}}, \mbox{Sayaka~Matsumoto~\orcidlink{0000-0002-4556-0744}}, \mbox{Tomohiro~Akiyama~\orcidlink{0000-0002-6720-8995}}, \mbox{Shimpei~Matsumoto~\orcidlink{0009-0005-6587-7852}}, \mbox{Katsuyuki~Kunida~\orcidlink{0000-0002-6346-2638}}, \mbox{Satoshi~Takahashi~\orcidlink{0000-0001-5573-7841}}, \mbox{Tokuro~Matsuo~\orcidlink{0009-0000-3551-8799}}, \mbox{Sachio~Hirokawa~\orcidlink{0000-0002-8050-1109}}, \mbox{Kazuya~Tsuruta~\orcidlink{0009-0004-7386-2290}}, \mbox{Ryosuke~Kozaki~\orcidlink{0000-0001-8428-9371}}, \mbox{Shuto~Nishida}, \mbox{Koma~Takamatsu~\orcidlink{0009-0001-8144-6797}}, \mbox{Ford~Lumban~Gaol~\orcidlink{0000-0002-5116-5708}}, and \mbox{Masao~Mori~\orcidlink{0000-0002-2883-8943}}%
\thanks{Kunihiko Takamatsu, Sayaka Matsumoto, and Masao Mori are with the Institute of Science Tokyo, Tokyo, Japan. K. Murakami is with The University of Tokyo, Tokyo, Japan. T. Akiyama is Specially Appointed Associate Professor, Uehiro Research Center for Japan Environmental Studies, Research Institute for Humanity and Nature, Japan; Associate Professor (co-appointed), Global Environmental Studies Program, Graduate University for Advanced Studies, SOKENDAI, Japan; Project Researcher, Department of Multi-Disciplinary Sciences, Graduate School of Arts and Sciences, The University of Tokyo, Japan; Visiting Professor, Graduate School of Information Technology, Kobe Institute of Computing, Japan; Visiting Professor, Miyazaki International University, Japan; and Visiting Researcher, Institute for Contemporary Chinese Studies, Aichi University, Japan. Shimpei Matsumoto is with Hiroshima Institute of Technology, Hiroshima, Japan. K. Kunida and T. Matsuo are with Fujita Health University, Aichi, Japan. S. Takahashi is with The University of Electro-Communications, Tokyo, Japan. S. Hirokawa is with Kyushu University, Fukuoka, Japan, and Monolithic Design Co., Ltd., Fukuoka, Japan. K. Tsuruta is with Monolithic Design Co., Ltd., Fukuoka, Japan. R. Kozaki and S. Nishida are with International Pacific University, Okayama, Japan. Koma Takamatsu is with Kwansei Gakuin University, Hyogo, Japan. F. Lumban Gaol is with Binus University, Jakarta, Indonesia. Corresponding author: Kunihiko Takamatsu (ORCID 0000-0003-4172-7612).}}

\begin{document}
\maketitle
\sloppy

\begin{abstract}
A Lean~4 proof of the Riemann Hypothesis was released on 20 November 2026 with 175,187 files and 17.4 million lines. Within a week it existed in three versions: the original release, a branch consolidated by volunteers (47,885 modules, 14.6 million lines), and a condensed rewriting that became the community's main line (802 files, 0.48 million lines, 2.8\% of the original). We ask whether the reduction came from pruning or from rewriting, and how far pruning alone could have gone. Using compiled indexes and declaration-level dependency graphs, we trace declarations across the versions. Consolidation kept the names: 92\% of the consolidated branch's real declarations have a counterpart of the same name in the original. The condensed version shares almost nothing with the consolidated branch: only 47 of its 26,018 real declarations in the dependency graph (0.18\%) match one by name and statement, and the nine full names common to all three versions are the fixed statements of the challenge. The declarations dropped were not the unused ones; their use profile is close to that of the whole branch. Removing every declaration of the consolidated branch that is unreachable from the five main theorems would cut its declaration lines by only 2.8\% (deleting whole declarations, without changing the others), whereas the condensed version has 2.7\% of the reachable lines. At the level of declaration text, the reduction is therefore almost entirely due to rewriting, not to pruning. The condensed version is also organised differently: it separates a library with no counted use of the proof-specific part, whereas the top-level namespaces of the consolidated branch all depend on one another. We discuss what this implies for checking and maintaining very large AI-written formal artifacts.
\end{abstract}

\section{Introduction}\label{sec:intro}

On 20 November 2026, Levent Alp{\"o}ge released, together with Claude, a Lean~4 development whose final theorems are Mathlib's statements of the Riemann Hypothesis and of the corresponding statements for Dirichlet $L$-functions~\cite{alpoge_archive}. On the next day the Lean FRO and the Mathlib Initiative reported that the statements are Mathlib's own and that the proof is accepted by the Lean kernel~\cite{leanfro_blog}. Within a week the development existed in three public versions of very different size, which we number as follows.
\begin{enumerate}
\item[(1)] The original release (commit \texttt{40e56bab02}): 175,187 files and 17,374,215 lines~\cite{alpoge_archive}.
\item[(2)] The volunteer-consolidated branch of \texttt{rohanarun/\allowbreak riemann-hypothesis} at commit \texttt{e497a66} (25 November 2026, 17:55 UTC), the last commit on that branch~\cite{repo_ra,pr120}: 47,885 modules, 14,583,569 lines and 871,784 real declarations, counted by the rule of our first paper~\cite{takamatsu2026unused}.
\item[(3)] The main branch of \texttt{leanprover-community/\allowbreak riemann-hypothesis} at commit \texttt{3b68810} (26 November 2026, 22:14 UTC)~\cite{repo_lc}: 802 files, 478,591 lines and 26,019 real declarations under the same rule.
\end{enumerate}
Version~(3) is based on the readable edition of \texttt{lsernau/grh-lean}~\cite{repo_grhlean}. The NOTICE of that edition, carried over into version~(3), states that it is ``a rewriting of Levent Alp{\"o}ge's Lean proof'', and the README of \texttt{lsernau/grh-lean} describes the repository as a shorter version ``produced in collaboration with my colleagues at Anthropic, and Claude''. Beyond these statements, how the rewriting was produced, and how the work was divided between people and Claude, is not public, and we make no claim about it. That version~(3) is a rewriting is therefore the authors' own statement, not a finding of ours. What has not been measured is how much of version~(2) survives in version~(3), and how small the development could have become by deletion alone.

\paragraph{The question.}
Version~(3) has 2.8\% of the lines of version~(1) and 1/30.5 of the lines of version~(2). A reduction of this size can arise in two ways. The declarations that the proof does not need can be pruned, leaving the rest as it was; or the proof can be written again, so that few of the old declarations remain at all. The two have different consequences for anyone who wants to read, check or teach the proof: a pruned version can be traced back, declaration by declaration, to the original, while a rewritten version is a new text that must be read on its own. We therefore ask two questions. Was version~(3) obtained from version~(2) by pruning or by rewriting? And how far could pruning alone have gone, that is, how many lines of version~(2) are reachable from its five main theorems? Existing measurements give the sizes of the versions~\cite{repo_rhshrink,repo_novusedge}, a skeleton of the original pruned by reachability~\cite{repo_foxfire}, reading notes~\cite{repo_engram}, and the number of constants of version~(1) reachable from the main theorems~\cite{zhuk9581,yth4377,schiller3750}; none of them traces individual declarations across the versions.

\paragraph{Contributions.}
We trace the declarations of the three versions by name and by normalized statement, and claim three results. For versions~(2) and~(3) we count real declarations from compiled indexes by the rule of our first paper; for version~(1), which we did not compile, we use the declarations found in its source.
\begin{enumerate}
\item \emph{Version~(3) is, at the level of declarations, almost entirely a rewriting of version~(2): almost none of its declarations match one of version~(2) in name and statement.} Only 47 declarations of version~(3) (0.18\%) have a strict counterpart in version~(2) (same full name; same short name and statement; or the same statement of at least 40 characters), and only 15 of these have a statement of substance (at least 20 characters). Nine of the 47 are the five main theorems and the four definitions of the statements, which were fixed as the task. In support of this, the declarations of version~(2) that did not survive are not concentrated among unused ones: their pattern of use (46.2\% used from another file, 51.9\% used only in their own file, 1.87\% with no recorded use) is close to that of version~(2) as a whole (47.9\%, 50.3\%, 1.84\%), and even among the 5,507 declarations used from at least 100 other modules, 17 survived. By contrast, the step from version~(1) to version~(2) kept names: 76.05\% of the named declarations in the source of version~(1) have a strict counterpart in version~(2).
\item \emph{We bound from below what pruning alone could reach.} From the declaration-level dependency graph of version~(2), the declarations reachable from the five main theorems number 826,654 of 871,784 (94.8\%) and occupy 10,432,860 lines of declaration spans (97.2\% of the 10,737,061 such lines of version~(2)), against 278,198 lines of declaration spans (478,591 lines in all) in version~(3). Version~(3) thus has 2.67\% of the declaration lines of this reachable part, about 1/37.5. The difference, 10,154,662 declaration lines, is the part of the reduction that pruning by reachability could not have produced. We do not claim that unreachable declarations can be deleted; the reachable lines are an estimate of how far deletion can go, not a deletion plan.
\item \emph{The reduction is in the number of declarations, not in their length.} Real declarations fell by a factor of 33.5, while the mean length of a declaration with an extracted statement stayed at 13.54 lines in version~(2) and 13.60 lines in version~(3). Dependence on external libraries (Mathlib and the Lean core) did not grow in proportion to the declarations: the module-level uses of external constants fell by a factor of 34.8, from 29.5 to 28.4 per real declaration, while the number of distinct external constants fell only from 69,633 to 32,220, and 82.6\% of those used by version~(3) had already been used by version~(2). The Mathlib commits of the two versions are 153 commits apart, so the names can be compared.
\end{enumerate}
A match of text is not a match of mathematics. We do not claim how many lemmas of version~(2) reappear in version~(3) in rewritten form; matches of short names alone are reported only as an upper indication.

\paragraph{Relation to the first paper.}
Our first paper~\cite{takamatsu2026unused} asked how far the report that a declaration of version~(2) is ``unused'' can be trusted, built a compiled forward and reverse index of version~(2) for that purpose, and defined the real declarations as the declarations left after compiler-generated auxiliary constants are attributed to their parents. The present paper reuses that index and that rule without repeating its results, and turns from a single version to the relation between versions. The approach is the same as in our earlier work on textbooks~\cite{takamatsu2026coverage}: collect the primary records, here every declaration of every version, and examine them directly rather than through aggregate sizes. In the terms of Eduinformatics~\cite{edu2018}, the question is one that arises with any teaching material that is revised: whether a new edition was made by cutting and correcting the old one, or by writing it again. A review of the first week of research on the release is given in~\cite{takamatsu2026review}.

\paragraph{Outline.}
Section~2 reviews the repositories that measured or reduced the versions, the reachability counts of version~(1), and the first paper. Section~3 describes the sources and commits of the three versions, the rule for real declarations, the normalization of statements, the correspondence classes, and how reachability was computed. Section~4 reports the sizes of the three versions, the correspondence from version~(2) to version~(3), the use of the declarations that did not survive, the lower bound for pruning, and the new parts of version~(3) together with their dependence on Mathlib. Section~5 discusses the two kinds of reduction, consolidation and rewriting, and the limits of tidying by deletion. Section~6 states the limitations, data and code availability, and the use of generative AI in this work.

\section{Background and related work}\label{sec:background}

\subsection{The release and the volunteer consolidation}\label{sec:bg-release}

On 20 November 2026 Alp{\"o}ge released, with Claude, a Lean~4 development of the Riemann Hypothesis and of the Generalized Riemann Hypothesis for Dirichlet $L$-functions~\cite{alpoge_archive}.
We call the first public version, preserved as commit \texttt{40e56bab02} on the \texttt{original} branch of \texttt{leanprover-community/\allowbreak riemann-hypothesis}~\cite{repo_lc}, version~(1); it has 175,187 files and 17,374,215 lines.
Within six days at least 89 preprints on the release appeared, none peer reviewed, of which nine were verification studies~\cite{takamatsu2026review}.
In parallel, about two hundred GitHub contributors began to rebuild, reduce and port the archive~\cite{takamatsu2026review,takamatsu2026unused}.
One restructuring, pull request \#120, reduced the number of \texttt{.lean} files from 149,396 to 47,884~\cite{pr120}; many later pull requests fixed the resulting build errors and removed declarations that appeared to be unused.
Merging on that branch, \texttt{rohanarun/\allowbreak riemann-hypothesis}~\cite{repo_ra}, stopped at commit \texttt{e497a66}, which we call version~(2).

Our first paper~\cite{takamatsu2026unused} studied version~(2) alone.
It built a forward and reverse index of the declarations from the compiled files, defined which compiled constants count as real declarations (as opposed to compiler-generated auxiliaries), and examined how reliable the volunteers' working rule ``no other file uses it, so it can be removed'' is against that index.
The present paper reuses that index and the same rule for real declarations, and refers to~\cite{takamatsu2026unused} for its results; here the question is not whether a single deletion is safe, but how the development changed between versions.
The first paper's approach in turn rests on a study of forward and reverse lookup on citation graphs of mathematics textbooks~\cite{takamatsu2026coverage}.

\subsection{The condensed rewrite}\label{sec:bg-rewrite}

On 25 November the \texttt{main} branch of \texttt{leanprover-community/\allowbreak riemann-hypothesis} was replaced by what its README calls ``a highly condensed version''; the progress made on the earlier development is kept on the \texttt{alpoge} branch~\cite{repo_lc}.
The new \texttt{main} is based on a copy of the readable edition (\texttt{grh/}) of \texttt{lsernau/grh-lean} at commit \texttt{72fead1}~\cite{repo_lc,repo_grhlean}.
We call \texttt{main} at commit \texttt{3b68810} version~(3).

The authors state the nature of this version themselves.
The README of \texttt{lsernau/grh-lean} says that the proof ``is based on Levent Alp{\"o}ge's, with Claude'' and that the repository ``is a shorter version, produced in collaboration with my colleagues at Anthropic, and Claude''; it also states that the proof ``has passed machine checks only'' and that ``No mathematician has read it through''~\cite{repo_grhlean}.
The NOTICE file of version~(3) says: ``This product is a rewriting of Levent Alp{\"o}ge's Lean proof'', and that \texttt{src/Statements.lean} and ``the names and types of the five theorems in \texttt{src/Main.lean} are taken from that proof unchanged''~\cite{repo_lc}.
That version~(3) is a rewriting is therefore not our finding.
What has not been measured is how much of version~(2) survives in version~(3) at the level of declarations, and how far deletion alone could have reduced version~(2).
We describe version~(3) only through these public statements; we do not know how the work was divided between people and Claude, and we make no claim about why the rewritten proof is shorter.

\subsection{Repositories that measure the versions}\label{sec:bg-repos}

Four public repositories measure or describe these versions; none of them is a paper.
\texttt{rh-shrink}~\cite{repo_rhshrink} reports the size of \texttt{leanprover-community/\allowbreak riemann-hypothesis} per commit on \texttt{main}, including \texttt{main} before its replacement on 25 November.
Its script counts lines, declarations, files and import statements with regular expressions over the file contents in git, without running Lean, and is rerun every 15 minutes.
\texttt{NovusEdge/rh-proof-notes}~\cite{repo_novusedge} measures, from source text and without building, the original archive and the readable and compact editions of \texttt{lsernau/grh-lean} (files, lines, code and comment lines, theorems and lemmas, repeated six-line blocks, and theorems whose name never appears again), and reports a build of the compact edition at commit \texttt{72fead1} that checks the axioms of the five main theorems.
Its comparisons are of totals; it does not trace which declarations of one version correspond to which of another, and it notes that its repetition measure is textual.
\texttt{Foxfire1st/rh-proof-skeleton}~\cite{repo_foxfire} is a pipeline that parses declarations, prunes them by reachability, and emits signature stubs grouped into eight condensed proof stages for a web explainer; it extracts a skeleton for readers rather than measuring how much of a version is reachable.
\texttt{engram-agents/riemann-explained}~\cite{repo_engram} is a set of reading notes, written by AI agents, that anchor each statement to the full name and file of the Lean declaration it describes; one note measures how much of the library stands behind each step of the proof. It does not compare versions, and we cite it for completeness.

Compared with these, our study counts declarations from the compiled files of version~(2) and version~(3) with one rule for real declarations, matches declarations across versions by name and by the normalised text of their statements (a textual match, not a test of mathematical identity), and, for version~(2), converts reachability from the main theorems into lines, so that a bound on the reduction that deletion alone could reach can be separated from the reduction obtained by rewriting.

\subsection{Reachability in the original release}\label{sec:bg-reach}

Three groups computed, for version~(1), the set of constants reachable from the five root theorems in the compiled environment: 1,478,222~\cite{zhuk9581}, 1,478,243~\cite{yth4377} and 1,478,219~\cite{schiller3750}.
Schiller reports two independent transitive traversals that agree~\cite{schiller3750}.
These counts refer to version~(1), are taken over the constants of the compiled environment rather than over real declarations in the sense of~\cite{takamatsu2026unused}, and are not expressed in lines.
They do not give the size of the part of version~(2) that the proof needs, which is what a lower bound on reduction by deletion requires.
Section~\ref{sec:method} describes how we compute reachability in version~(2) at the level of real declarations, and Section~\ref{sec:results} reports the result.
As in~\cite{takamatsu2026unused}, we do not read ``unreachable'' as ``deletable'': a declaration that the main theorems do not need can still be used by another declaration, so its removal can break the build.
The size of the reachable part is used only as a lower bound on the size that deletion alone could reach, not as a list of declarations that could be removed.

\subsection{Clone detection, slicing and the maintenance of proof libraries}\label{sec:bg-se}

Matching declarations across versions by name and normalised text is close to what software engineering calls clone detection: our strict match corresponds to exact and renamed (type-1 and type-2) clones at the level of declarations, and category~E to near-miss clones~\cite{roy2009clones}. Studies of clone genealogies and clone evolution follow such matches across the versions of a code base~\cite{kim2005genealogies,gode2013evolution}; we use the same idea for three versions of a proof, but at the granularity of declarations and without detecting near-miss clones beyond category~E. Reachability from the main theorems is a coarse, declaration-level form of the dependence that program slicing computes~\cite{weiser1984slicing}. On the side of formal proofs, the engineering of large verified developments, including their maintenance and refactoring, has been surveyed~\cite{ringer2019qed}, and the design and maintenance of Mathlib, on which all three versions build, have been described by its developers~\cite{mathlib2020,vandoorn2020maintaining}. These works concern code and libraries written and maintained by people over years; the versions studied here were produced within a week, versions~(1) and~(3) with substantial use of an AI system.

\subsection{Eduinformatics framing}\label{sec:bg-edu}

The study follows the practice of Eduinformatics, a field proposed in 2018 that integrates education and informatics~\cite{edu2018}: to collect primary records exhaustively and analyse them directly, rather than through aggregated indicators that can hide what the records contain~\cite{takamatsu2026coverage,takamatsu2026unused}.
Here the primary records are all real declarations of each version, and the aggregate that can mislead is the ratio of total sizes.
The question also has a counterpart in the revision of teaching materials: when a textbook or a course becomes much shorter, was it pruned and corrected, or written anew?
Totals cannot tell these apart; tracing the individual items across editions can.
We do not extend the framing further.

\section{Data and method}\label{sec:method}

This section states which three versions we compare, how declarations are counted in each, how declarations are matched across versions, and how reachability from the main theorems is computed. The index and the counting rule are those of our first paper~\cite{takamatsu2026unused}; we summarise them only as far as needed here and do not repeat its results. No Lean source text of versions~(1) and~(2), which carry no licence, is redistributed; we report names, categories and counts only.

\subsection{Three versions}\label{sec:method:versions}

\begin{itemize}
\item \textbf{Version (1)}, the original release of 20 November 2026~\cite{alpoge_archive}, taken as commit \texttt{40e56bab02} (``Mirror the submission exactly as published'') of \texttt{rohanarun/\allowbreak riemann-hypothesis}~\cite{repo_ra}: 175,187 \texttt{.lean} files and 17,374,215 lines.
\item \textbf{Version (2)}, the volunteer-consolidated branch \texttt{opt/collapse-and-consolidate-4.36} of the same repository at commit \texttt{e497a66} (25 November 2026), the version studied in~\cite{takamatsu2026unused}: 47,885 modules and 14,583,569 lines, built with Lean v4.36.0 and Mathlib \texttt{fc4f59e}.
\item \textbf{Version (3)}, the \texttt{main} branch of \texttt{leanprover-community/\allowbreak riemann-hypothesis}~\cite{repo_lc} at commit \texttt{3b68810} (26 November 2026, 22:14 UTC): 802 \texttt{.lean} files under \texttt{src/} and 478,591 lines, built with Lean v4.37.0-rc1 and Mathlib \texttt{31ce06e}. Its README states that the branch is based on a copy of the readable edition of \texttt{lsernau/grh-lean}~\cite{repo_grhlean}, and its NOTICE states that it ``is a rewriting of Levent Alp\"oge's Lean proof'', with \texttt{Statements.lean} and the names and types of the five theorems in \texttt{Main.lean} taken over unchanged.
\end{itemize}
Version~(3) is still changing; all of its numbers refer to commit \texttt{3b68810}. Lines are counted as newline characters in all \texttt{.lean} files under \texttt{src/} (checked twice, with \texttt{wc -l} and with a separate script); we also report non-blank lines. The two Mathlib commits are close: \texttt{31ce06e} is a descendant of \texttt{fc4f59e}, 153 commits later.

\subsection{Compiled indexes and real declarations}\label{sec:method:index}

For versions~(2) and~(3) we use indexes read from the compiled \texttt{.olean} files, not from source text~\cite{takamatsu2026unused}. For each module the index has three kinds of rows: \textbf{D} (module $m$ declares name $n$), \textbf{U} (module $m$ uses name $n$ declared in another module) and \textbf{S} (declaration $a$ uses declaration $b$ in the same module, $a\neq b$). For version~(2) we reuse the index of~\cite{takamatsu2026unused}, which covers the 47,780 of 47,885 modules that compile locally; the 105 uncovered modules declare nothing of their own except three definitions in the statement file of version~(2). For version~(3) we built the same index for all 802 modules in the same form (nine runs, all of which completed).

Each index contains many constants generated by the elaborator (proof, simp and match auxiliaries, equation lemmas, recursors and constructor helpers, notation helpers, and the prefixed auxiliaries of private declarations). As in~\cite{takamatsu2026unused}, every name is mapped to its \emph{root}, the prefix before the first auxiliary component, and a declaration is \emph{real} if it is its own root; a use of an auxiliary is counted as a use of its root. We apply the same rule, with the same code, to both versions. Version~(2) has 871,784 real declarations.

For version~(3) we also extracted the declaration-level dependency graph: for every project constant, an \textbf{E} row from the constant to each project constant used in its type or value, read from the compiled environment (61,985 D rows and 682,995 E rows). Version~(3) has 26,019 real declarations in the D/U/S index, the count we report as its size throughout, and 26,018 in the D/E graph. The two differ in two compiler-generated congruence lemmas (\texttt{*.hcongr\_N}): \texttt{Subgroup.toGroup.hcongr\_3} is a real declaration among the D rows of the index but absent from the D rows of the graph, and \texttt{Pi.single.hcongr\_7} has no use from another module in the index but one in the graph; all other real declarations agree. Counts of use in version~(3) are taken from the D/U/S index, so that they are computed in the same way as in version~(2); matching and reachability use the D/E graph. A third extraction with a modified program also writes rows from project declarations to external constants; we use it, together with the U rows of both versions, only to compare the use of Mathlib in the results. An external constant is a name in a U row that is not declared in any module of the project, according to the D rows.

\paragraph{Use classes.}
Following~\cite{takamatsu2026unused}, $U(n)$ is the number of other modules that use the real declaration $n$, after folding auxiliaries. A real declaration with $U(n)=0$ is a \emph{false zero} if some other declaration of its own file uses it (an S row whose using side has a root different from $n$), and a \emph{true zero} otherwise. For the comparison of retained and lost declarations we group $U$ into the classes $0$, $1$, $2$--$9$, $10$--$99$ and ${\ge}100$.

\subsection{Statements and normalisation}\label{sec:method:stmt}

Statements are extracted from source by one rule for all versions. A declaration line that begins in column~0 opens a declaration; its statement is the text after the name up to the first \texttt{:=}, at most 700 characters and at most 15 lines from the declaration line. Statements were attached to 792,882 of the real declarations of version~(2) and 20,454 of version~(3). Declarations without a statement line (structure fields and constructors, anonymous instances, macro-generated declarations, and declarations that do not begin in column~0) enter only the name-based categories below. Before comparison a statement is normalised: attributes \texttt{@[\ldots]}, modifiers such as \texttt{private}, the declaration keyword and the declaration name are removed, and runs of white space are collapsed to one space. Full names are compared after removing the prefix \texttt{\_private.\textit{M}.0.}; the \emph{short name} is the last component of the full name.

Normalised statements are compared as text. Equal text does not imply the same mathematical content, since \texttt{open}, \texttt{variable} and \texttt{namespace} contexts can differ; conversely, the same lemma restated with renamed types or reordered arguments is not found. The matching therefore detects declarations that were carried over in text, not mathematical identity.

\subsection{Matching categories}\label{sec:method:match}

Each real declaration of one version is placed in the first of the categories in Table~\ref{tab:categories} that applies with respect to the other version; lower categories are not examined once a higher one applies. Matching is between sets: a declaration of one version may match several of the other, so the two sides of a pair of versions do not have equal counts. We call A$+$B$+$C40 a \emph{strict} match. Category~C consists mostly of generic types such as \texttt{: Prop}, \texttt{: Type} or \texttt{:}~$\mathbb{N}$, which coincide by chance; it is reported only to bound the matching and is not counted as correspondence. Category~E is an upper indication of declarations that may have been rewritten under the same name; it is not counted as correspondence either.

\begin{table}[t]
\centering
\caption{Matching categories, applied in this order.}
\label{tab:categories}
\begin{tabularx}{\linewidth}{@{}lX@{}}
\toprule
Category & Condition with respect to the other version \\
\midrule
A & same full name \\
B & same short name and same normalised statement \\
C40 & same normalised statement of at least 40 characters, different name \\
C & same normalised statement of fewer than 40 characters, different name \\
E & different statement; same short name of at least 15 characters \\
N & none of the above \\
\bottomrule
\end{tabularx}
\end{table}

The categories were computed by one program and recomputed by an independent program that normalises statements differently and does not read the first program's output; the two agreed on every category count for versions~(2) and~(3) (the second program reports C and C40 together, and its totals equal their sums). A random sample of 300 version-(3) declarations in category~N with statements of at least 40 characters was also searched, as substrings of the first 60 characters of the statement without the name, in the raw statements of version~(2).

\subsection{Version (1) without compilation}\label{sec:method:v1}

We did not compile version~(1); with 3.7 times the files of version~(2), whose full build took about ten hours, it could not be built within the time of this study. Its declarations are therefore \emph{source-parsed}: named declarations found by the source parser used for the partition comparison in~\cite{takamatsu2026unused}, 1,057,984 in total (1,051,775 distinct full names, since files that do not import each other may reuse a name), plus 3,547 anonymous instances that cannot be matched by name. An independent scanner written for this study found 1,057,878; 418 and 524 names (under 0.05\% each) were found by only one of the two. Statements were attached to 1,042,271 of the parsed declarations by the extraction rule above. Version~(1) is matched against the real declarations of versions~(2) and~(3) with the same categories. On the side of version~(2) we count separately, as \textbf{Ap}, declarations without a statement line (fields, constructors, generated names) that are not in A but whose parent name is declared in version~(1). Counts for version~(1) are counts of source-parsed declarations and are not directly comparable with counts of real declarations; source-parsed declarations of version~(2) number 805,873 against its 871,784 real declarations, the difference being mainly structure fields and constructors, which have no name in the source.

\subsection{Reachability and the pruning bound}\label{sec:method:reach}

Reachability is taken from the five theorems of \texttt{Main.lean} (\texttt{riemannHypothesis}, \texttt{riemannZeta\_eq\_zero\_re}, \texttt{LFunction\_eq\_zero\_re}, \texttt{LFunction\_eq\_zero\_re\_of\_isPrimitive}, \texttt{LFunction\_ne\_zero\_of\_half\_lt\_re}) by breadth-first search over the declaration-level E rows. An auxiliary constant and its root are identified, so that a declaration reached only through its own proof or simp auxiliary is counted as reached. For version~(3) the search runs on the D/E graph above and was checked by a second, independent implementation.

The earlier counts of what is reachable from the five main theorems in version~(1), 1,478,222~\cite{zhuk9581}, 1,478,243~\cite{yth4377} and 1,478,219~\cite{schiller3750}, were obtained on the compiled environment of the original release and are not counts of real declarations in the sense used here. A skeleton pruned by reachability has also been extracted by others~\cite{repo_foxfire}. We need instead the number of \emph{real} declarations of version~(2) that are reachable, and the lines of source they occupy. The existing index of version~(2) cannot give this, because its cross-file rows (U) go from a using module to a used name, not from declaration to declaration; it yields only a module-level closure, which is an upper bound. We therefore extract declaration-level E rows for version~(2) from its compiled files. The 47,780 indexed modules are split into 120 chunks of 400; for each chunk a small Lean program imports the chunk's modules (with their dependencies) and writes, for every constant declared in those modules, the project constants that its type and value use. Each declaration is assigned to its module by the existing D rows. The extraction is accepted only if its E rows, folded to modules, reproduce the existing cross-file use rows. It produced 16,035,877 E rows. On 1,346 randomly sampled modules, the E rows folded to modules agree with the existing cross-file use index to within 0.028\% (uses in the index missing from the folded graph) and 0.080\% (uses in the folded graph missing from the index). The reachability of version~(2) reported below is therefore an exact count on the declaration-level graph, not the module-level upper bound.

To turn reachable declarations into lines, each declaration with a statement line is assigned its source span, from its declaration line to the next line that begins in column~0. Summed over declarations with a statement, these spans give 13.54 lines per declaration in version~(2) and 13.60 in version~(3). The lines of the reachable real declarations of version~(2) are an estimate of the size that pruning alone could reach; we compare this with the size of version~(3). Reachable declarations without a statement line contribute no lines, and lines outside every declaration span (imports, comments, \texttt{namespace} and \texttt{variable} lines) are not counted; the totals are therefore \emph{declaration lines} (10,737,061 in version~(2), 278,198 in version~(3)), smaller than the file line counts (14,583,569 and 478,591), and the two versions are compared on declaration lines. We do not claim that unreachable declarations are deletable: a declaration that the proof does not need can still be used by another declaration, in its own or in another file~\cite{takamatsu2026unused}, and a name used in the source (for example in a \texttt{simp} or \texttt{rw} call, or through an attribute) need not leave any trace in the compiled terms from which reachability is computed. The lines of the reachable part are used only as a lower bound on the size to which deletion alone, without rewriting, could reduce version~(2).

\section{Results}\label{sec:results}

Throughout this section, version~(1) is the original release (commit \texttt{40e56bab02})~\cite{alpoge_archive}, version~(2) is the volunteer-consolidated branch (commit \texttt{e497a66})~\cite{repo_ra}, and version~(3) is the rewriting at \texttt{leanprover-community/\allowbreak riemann-hypothesis} (commit \texttt{3b68810})~\cite{repo_lc,repo_grhlean}.
For versions~(2) and~(3), ``declarations'' are real declarations counted from the compiled index with the rule of the first paper~\cite{takamatsu2026unused}; we did not build version~(1), so its declarations are named declarations parsed from the source.
Textual matches of names or statements are not claims of mathematical identity.

\subsection{Sizes of the three versions}\label{sec:res-size}

Table~\ref{tab:sizes} gives the sizes.
Version~(3) has 2.75\% of the lines of version~(1) and 3.28\% of the lines of version~(2).
From version~(2) to version~(3), the number of real declarations fell by a factor of 33.5 (871{,}784 to 26{,}019) and the number of lines by a factor of 30.5, so the number of lines per real declaration changed little (16.73 to 18.39).
Restricted to the declarations whose statement could be extracted from the source, the span of a declaration (from its header line to the next line starting in column zero) averaged 13.54 lines in version~(2) and 13.60 lines in version~(3).
What shrank was the number of declarations, not their length.

\begin{table*}[t]
\centering
\caption{Sizes of the three versions. Declarations of version~(1) are named declarations parsed from the source (we did not build version~(1)); those of versions~(2) and~(3) are real declarations from the compiled index~\cite{takamatsu2026unused}. Source-parsed counts are also given for versions~(2) and~(3), with the same parser.}
\label{tab:sizes}
\footnotesize
\setlength{\tabcolsep}{3pt}
\begin{tabular}{@{}lrrr@{}}
\toprule
 & (1) \texttt{40e56bab02} & (2) \texttt{e497a66} & (3) \texttt{3b68810} \\
\midrule
\texttt{.lean} files & 175{,}187 & 47{,}885 & 802 \\
Lines & 17{,}374{,}215 & 14{,}583{,}569 & 478{,}591 \\
Non-blank lines & 15{,}531{,}759 & 12{,}915{,}887 & 433{,}349 \\
Named declarations (source) & 1{,}057{,}984 & 805{,}873 & 22{,}309 \\
Real declarations (compiled) & --- & 871{,}784 & 26{,}019 \\
Lines per named declaration & 16.42 & 18.10 & 21.45 \\
Lines per real declaration & --- & 16.73 & 18.39 \\
Declaration span, mean lines$^{\ast}$ & --- & 13.54 & 13.60 \\
\bottomrule
\end{tabular}

\smallskip
\raggedright
$^{\ast}$Over declarations with an extracted statement (792{,}882 in version~(2), 20{,}454 in version~(3)).
\end{table*}

The count 26{,}019 for version~(3) uses the same index format and rule as version~(2).
The declaration-level mapping and the reachability below use the dependency graph of version~(3), in which real declarations number 26{,}018 (Section~\ref{sec:method:index}).

\subsection{Mapping between versions}\label{sec:res-map}

\begin{figure}[t]
\centering
\includegraphics[width=\columnwidth]{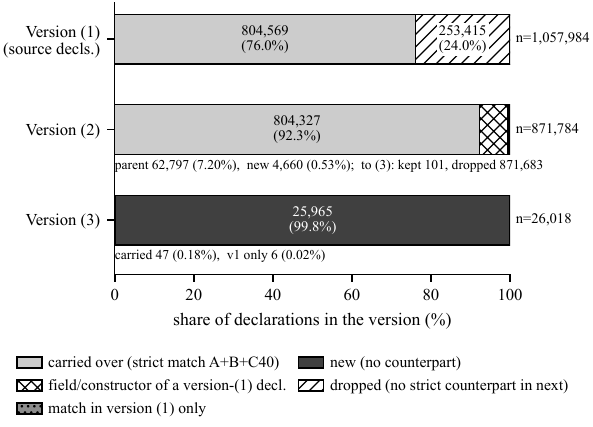}
\caption{Declarations of the three versions. (a) Linear axis; (b) logarithmic axis. Grey: strict counterpart (A+B+C40) in the adjacent version; light grey: fields and constructors of version~(1) declarations (Ap); hatched: no strict counterpart in the next version; dotted: no strict counterpart in the previous version. Version~(1) is counted as named source declarations (we did not build it); versions~(2) and~(3) as real declarations from the compiled index. The flow from version~(2) to version~(3) (101 declarations to 47) is too thin to see on the linear axis and is marked by a line.}
\label{fig:band}
\end{figure}

Table~\ref{tab:map} gives the categories of the mapping, and Fig.~\ref{fig:band} shows the three versions side by side.
The categories are assigned in the order A (same full name), B (same short name and same normalized statement), C40 (same statement of at least 40 characters, other name), C (same statement under 40 characters, mostly generic types such as \texttt{: Prop}), E (same short name of at least 15 characters only), N (none).
We call A+B+C40 a \emph{strict} counterpart; C and E are reported as upper-bound indications only.
Matching is between sets and is not one-to-one, so the two sides of a pair of versions need not have equal counts.

\begin{table*}[t]
\centering
\caption{Mapping categories between versions. Counts with shares of each version. Version~(1) is counted as named source declarations, versions~(2) and~(3) as real declarations (26{,}018 for version~(3); see Section~\ref{sec:res-size}). Matching is set-to-set, not one-to-one.}
\label{tab:map}
\footnotesize
\begin{tabular}{@{}lrrrrrrrr@{}}
\toprule
 & \multicolumn{4}{c}{(1)$\leftrightarrow$(2)} & \multicolumn{4}{c}{(2)$\leftrightarrow$(3)} \\
\cmidrule(lr){2-5}\cmidrule(l){6-9}
Category & (1) & \% & (2) & \% & (2) & \% & (3) & \% \\
\midrule
A same full name & 804{,}561 & 76.05 & 804{,}319 & 92.26 & 9 & 0.001 & 9 & 0.035 \\
B short name + statement & 3 & 0.0003 & 3 & 0.0003 & 76 & 0.009 & 36 & 0.138 \\
C40 statement ($\ge$40 chars) & 5 & 0.0005 & 5 & 0.0006 & 16 & 0.002 & 2 & 0.008 \\
\textbf{Strict (A+B+C40)} & \textbf{804{,}569} & \textbf{76.05} & \textbf{804{,}327} & \textbf{92.26} & \textbf{101} & \textbf{0.012} & \textbf{47} & \textbf{0.181} \\
Ap parent name in (1) & --- & --- & 62{,}797 & 7.20 & --- & --- & --- & --- \\
C generic statement only & 19{,}387 & 1.83 & 2 & 0.0002 & 30{,}270 & 3.47 & 489 & 1.88 \\
E short name only ($\ge$15 chars) & 6{,}858 & 0.65 & 158 & 0.02 & 1{,}015 & 0.12 & 385 & 1.48 \\
N none & 227{,}170 & 21.47 & 4{,}500 & 0.52 & 840{,}398 & 96.40 & 25{,}097 & 96.46 \\
\midrule
Total & 1{,}057{,}984 & & 871{,}784 & & 871{,}784 & & 26{,}018 & \\
\bottomrule
\end{tabular}
\end{table*}

\paragraph{Version (1) to version (2)}
The volunteer consolidation kept names.
Of the 1{,}057{,}984 named source declarations of version~(1), 804{,}569 (76.05\%) have a strict counterpart in version~(2), almost all by identical full name.
In the other direction, 804{,}327 (92.26\%) of the real declarations of version~(2) have a strict counterpart in version~(1), and 867{,}124 (99.47\%) when the fields and constructors of version~(1) structures (category Ap) are added; only 4{,}660 (0.53\%) appeared newly.
Renaming was rare (B and C40 together: 8 declarations).
Of the 253{,}415 declarations of version~(1) without a strict counterpart in version~(2), 150{,}488 lay in the 54{,}066 files of which no declaration was kept, and only 6{,}858 share a short name with a declaration of version~(2) (category E).
The reduction from version~(1) to version~(2) was therefore a removal of declarations and files, with the surviving declarations carried over under their names.

\paragraph{Version (2) to version (3)}
The mapping supports C1: at the level of declarations, version~(3) is almost entirely rewritten. Because matching is between sets, the 47 declarations of version~(3) with a strict counterpart correspond to 101 declarations of version~(2) (Table~\ref{tab:survival}).
Only 47 declarations of version~(3) (0.181\%) have a strict counterpart in version~(2).
The 9 declarations in category A are the five theorems of \texttt{Main} and four definitions of \texttt{Statements}, the part fixed as the statement of the challenge; their statements are also textually identical.
Of the 36 declarations of version~(3) in category B, 32 have statements shorter than 20 characters (such as \texttt{: Type} or \texttt{:}~$\mathbb{N}$); counting statements of at least 20 characters, 15 declarations of version~(3) have an informative match.
Version~(3) has 25{,}097 declarations (96.46\%) in category N, and 25{,}586 (98.34\%) in N or C; adding the 385 short-name-only matches of category E gives 25{,}971 (99.82\%).
Looser matching does not change the picture: of the 14{,}901 statements of version~(3) with at least 60 characters, 59 (0.40\%) share their first 60 characters with a statement of version~(2).
Comparing version~(1) and version~(3) directly gives 53 strict counterparts on the side of version~(3) (0.204\%); 47 of them are the 47 found through version~(2), and the remaining 6 are short definitions whose statement had changed in version~(2).
Of the 26{,}018 declarations of version~(3), 25{,}965 (99.80\%) have no strict counterpart in either earlier version.
That version~(3) is a rewriting is stated in the NOTICE of the repository it is based on~\cite{repo_grhlean}; these counts measure the extent of the rewriting at the level of declarations, not how it was produced.

\subsection{How the dropped declarations were used}\label{sec:res-use}

If version~(3) had been obtained by pruning, one would expect the declarations of version~(2) that disappeared to be concentrated among those that version~(2) itself did not use.
They are not.
This result is a consequence of C1 and is reported only in support of it.

\begin{figure}[t]
\centering
\includegraphics[width=\linewidth]{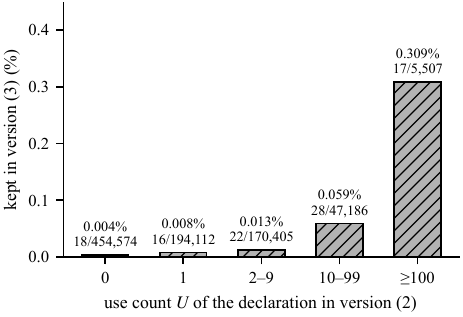}
\caption{Share of declarations of version~(2) with a strict counterpart in version~(3), by the number $U$ of other modules using the declaration in version~(2). Even among the most used declarations ($U\ge 100$) only 17 of 5,507 survive.}
\label{fig:survival}
\end{figure}

Table~\ref{tab:survival} and Fig.~\ref{fig:survival} give the share of declarations of version~(2) with a strict counterpart in version~(3), by the number $U$ of other modules that use the declaration in version~(2).
The share rises with $U$, but even among the 5{,}507 declarations used by at least 100 other modules only 17 (0.31\%) have a strict counterpart; many of those kept are small definitions of parameters or notation, with statements such as \texttt{:}~$\mathbb{N}$.

\begin{table}[t]
\centering
\caption{Declarations of version~(2) with a strict counterpart (A+B+C40) in version~(3), by the number $U$ of other modules that use them in version~(2).}
\label{tab:survival}
\footnotesize
\begin{tabular}{@{}lrrr@{}}
\toprule
$U$ & Declarations & Strict counterpart & \% \\
\midrule
0 & 454{,}574 & 18 & 0.0040 \\
1 & 194{,}112 & 16 & 0.0082 \\
2--9 & 170{,}405 & 22 & 0.0129 \\
10--99 & 47{,}186 & 28 & 0.0593 \\
$\ge$100 & 5{,}507 & 17 & 0.3087 \\
\midrule
Total & 871{,}784 & 101 & 0.0116 \\
\bottomrule
\end{tabular}
\end{table}

Table~\ref{tab:dropped} compares the 840{,}398 declarations of version~(2) in category N with the whole of version~(2).
The usage classes (used from another file; ``false zero'', unused from other files but used in the same file; ``true zero'') follow the first paper~\cite{takamatsu2026unused}; for the whole of version~(2), the counts of real declarations, of false zeros and of true zeros agree with that paper.
On every indicator, the dropped declarations have nearly the same shares as the whole: 388{,}632 of them were used from another file and 49{,}549 by at least 10 other modules.
The disappearance of declarations of version~(2) shows little selection by recorded use. (Recorded use is not reachability: a declaration used by other modules can still be unreachable from the main theorems; reachability is treated in Section~\ref{sec:res-reach}.)

\begin{table}[t]
\centering
\caption{Declarations of version~(2) without any counterpart in version~(3) (category N) compared with the whole of version~(2). Usage classes as in~\cite{takamatsu2026unused}.}
\label{tab:dropped}
\footnotesize
\setlength{\tabcolsep}{3pt}
\begin{tabular}{@{}lrr@{}}
\toprule
 & Dropped (N) & Whole of (2) \\
\midrule
Declarations & 840{,}398 & 871{,}784 \\
Used from another file & 388{,}632 (46.2\%) & 417{,}210 (47.9\%) \\
False zero & 436{,}041 (51.9\%) & 438{,}545 (50.3\%) \\
True zero & 15{,}725 (1.87\%) & 16{,}029 (1.84\%) \\
Used by $\ge$10 modules & 49{,}549 (5.9\%) & 52{,}693 (6.0\%) \\
Thin wrapper (heuristic) & 98{,}302 (11.7\%) & 101{,}432 (11.6\%) \\
Duplicated statement & 79{,}534 (9.5\%) & 79{,}644 (9.1\%) \\
\texttt{private} & 62{,}665 (7.5\%) & 62{,}730 (7.2\%) \\
\bottomrule
\end{tabular}

\smallskip
\raggedright
Thin wrapper: a body consisting of one application or one tactic step, at most 160 characters, judged from the source text only. Duplicated statement: a normalized statement of at least 40 characters shared by two or more real declarations of version~(2).
\end{table}

\subsection{Lower bound reachable by pruning alone}\label{sec:res-reach}

The mapping shows that version~(3) was not obtained by pruning version~(2); it does not show how far pruning alone could have gone.
To estimate this, we count the real declarations of version~(2) reachable from the five theorems of \texttt{Main} and the lines they occupy (Section~\ref{sec:method}).
The five theorems of \texttt{Main} in version~(2) are the same five as in version~(3).
The declaration-level graph of version~(2) has 16{,}035{,}877 E rows; folded to modules, it agrees with the existing cross-file use index to within 0.028\% and 0.080\% in the two directions on 1{,}346 randomly sampled modules (Section~\ref{sec:method:reach}).
Of the 871{,}784 real declarations of version~(2), 826{,}654 (94.8\%) are reachable from the five theorems and 45{,}130 are not; 46{,}960 of the 47{,}780 modules contain at least one reachable declaration.

\begin{figure}[t]
\centering
\includegraphics[width=\columnwidth]{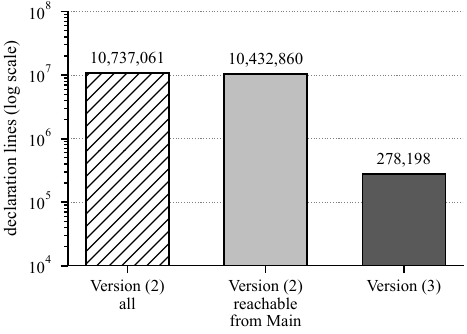}
\caption{Declaration lines of the whole of version~(2), of its part reachable from the five theorems of \texttt{Main}, and of version~(3), on a logarithmic axis. The step from the whole of version~(2) to its reachable part (2.83\% of declaration lines) is what pruning by reachability could remove in this compiled graph; the step from the reachable part to version~(3) (to 2.67\%) is not obtainable by pruning alone.}
\label{fig:prune}
\end{figure}

Table~\ref{tab:prune} and Fig.~\ref{fig:prune} give the lines.
The reachable declarations of version~(2) occupy 10{,}432{,}860 of its 10{,}737{,}061 declaration lines (97.2\%); removing every unreachable declaration would remove only 304{,}201 declaration lines, 2.83\% of the total.
Version~(3) has 278{,}198 declaration lines (478{,}591 lines in all), 2.67\% of the declaration lines of the reachable part of version~(2), or about 1/37.5.
In this compiled graph, therefore, pruning by reachability cannot bring version~(2) below about 10.4 million declaration lines, and the further 10{,}154{,}662 declaration lines (97.3\% of the reachable part) by which version~(3) is smaller are a reduction that pruning alone could not have produced.
Almost all of the reduction from version~(2) to version~(3) is thus attributable to rewriting, not to deletion.

\begin{table}[t]
\centering
\caption{Real declarations and declaration lines of version~(2), in all and reachable from the five theorems of \texttt{Main}, and of version~(3). Declaration lines are the sums of declaration spans (Section~\ref{sec:method:reach}). Share: declaration lines as a percentage of all declaration lines of version~(2); the 278{,}198 declaration lines of version~(3) are 2.59\% of all declaration lines of version~(2) and 2.67\% of its reachable part. The reachable part is a lower bound on the size that deletion by reachability could reach in this compiled graph, not a list of declarations that can be kept or deleted.}
\label{tab:prune}
\footnotesize
\setlength{\tabcolsep}{3pt}
\begin{tabular}{@{}lrrr@{}}
\toprule
 & Real declarations & Declaration lines & Share (\%) \\
\midrule
Version~(2), all & 871{,}784 & 10{,}737{,}061 & 100.0 \\
Version~(2), reachable & 826{,}654 & 10{,}432{,}860 & 97.2 \\
Version~(3) & 26{,}019 & 278{,}198 & 2.59 \\
\bottomrule
\end{tabular}
\end{table}

For comparison, in version~(3) itself 25{,}613 of the 26{,}018 real declarations (98.44\%) are reachable from \texttt{Main}; among the 25{,}097 declarations of version~(3) in category N, 24{,}699 (98.4\%) are reachable.
For version~(1), three groups reported 1{,}478{,}222, 1{,}478{,}243 and 1{,}478{,}219 constants reachable from the five root theorems~\cite{zhuk9581,yth4377,schiller3750}; these counts are taken over the compiled environment of version~(1), not in our unit of real declarations, and are not compared directly.

The reachable part is a lower bound on what pruning by reachability would leave, not a statement that every unreachable declaration can be deleted.
In version~(3), 322 of the 326 real declarations with no use at all in the compiled index are unreachable; of the 114 of these 326 for which deletion attempts gave a result, 111 failed, 69 of them with the deleted name reported as an unknown identifier or constant (the other 42 failed without naming it, so side effects of the deletion cannot be fully excluded for them).
A use that does not appear in the compiled terms, as described in the first paper~\cite{takamatsu2026unused}, can make an unreachable declaration necessary for the build.

\subsection{New material in version (3) and dependence on external libraries}\label{sec:res-ext}

The rewritten material is spread over the whole of version~(3).
In each of its twelve largest areas (by the first two components of the module path), the share of declarations without any counterpart (category N) lies between 93.5\% (\texttt{Lib.General}, 1{,}816 of 1{,}942) and 98.4\% (\texttt{New.Congruences}, 1{,}684 of 1{,}712); the largest area, \texttt{New.Spanning}, has 2{,}958 of 3{,}028 in category N.
The file \texttt{ATTRIBUTION.md} of version~(3) has 126 lines naming declarations of version~(3) that are stated to be modelled on Mathlib declarations; by that file's own account, part of the new material took Mathlib, rather than version~(2), as its model.

Table~\ref{tab:ext} compares the use of external constants, that is, names used by a module that are declared in no module of the project.
The comparison is by name (and by module-level use), not by meaning; it is possible because the two Mathlib versions are close: version~(2) uses Mathlib \texttt{fc4f59e} (Lean v4.36.0) and version~(3) uses Mathlib \texttt{31ce06e} (Lean v4.37.0-rc1), a descendant 153 commits later; only 16 of the 69{,}633 external names of version~(2) (0.023\%) are absent from the environment of version~(3).
Counts are at the module level (pairs of module and external constant), the only level available for both versions.

\begin{table}[t]
\centering
\caption{External constants used by versions~(2) and~(3), counted at the module level. ``Uses'': pairs (module, external constant). Library assignment by lookup in the environment of version~(3).}
\label{tab:ext}
\footnotesize
\setlength{\tabcolsep}{3pt}
\begin{tabular}{@{}lrrr@{}}
\toprule
 & (2) & (3) & (2)/(3) \\
\midrule
Real declarations & 871{,}784 & 26{,}019 & 33.5 \\
Distinct external constants & 69{,}633 & 32{,}220 & 2.16 \\
\quad Mathlib & 66{,}292 & 29{,}594 & \\
\quad Lean core (Init/Lean/Std) & 3{,}258 & 2{,}192 & \\
\quad Batteries / Aesop & 66 / 1 & 21 / 1 & \\
\quad TauCeti & --- & 412 & \\
\quad Not in env.\ of (3) & 16 & 0 & \\
Uses (module $\times$ constant) & 25{,}707{,}535 & 739{,}242 & 34.8 \\
\quad Mathlib share of uses & 79.5\% & 79.3\% & \\
Uses per real declaration & 29.5 & 28.4 & \\
Uses per line & 1.76 & 1.54 & \\
Distinct constants per real decl. & 0.080 & 1.24 & \\
\midrule
Used in both versions & \multicolumn{2}{c}{26{,}599} & \\
Used only in (2) / only in (3) & 43{,}034 & 5{,}621 & \\
\bottomrule
\end{tabular}
\end{table}

The uses fell by a factor of 34.8, close to the factor of 33.5 for real declarations, so the density of external dependence per declaration (29.5 to 28.4 uses) and per line (1.76 to 1.54) changed little, and the Mathlib share of uses stayed at about 79\%.
The number of distinct external constants fell only by a factor of 2.16, and 26{,}599 (82.6\%) of the external constants used by version~(3) were also used by version~(2).
Version~(3) uses 5{,}621 external constants that version~(2) did not use, 4{,}727 of them from Mathlib and 412 from TauCeti, a library that version~(2) does not depend on.
We therefore do not say that version~(3) depends more on Mathlib: its uses of external constants decreased in proportion to its declarations, while the number of distinct external constants used is still 0.46 times that of version~(2), and 38.2\% of the distinct external constants used by version~(2) are also used by version~(3) (C4).
Whether the Mathlib constants used only by version~(3) already existed in the Mathlib of version~(2) was not checked.

\subsection{Organisation of the versions}\label{sec:res-org}

\begin{figure*}[t]
\centering
\includegraphics[width=\textwidth]{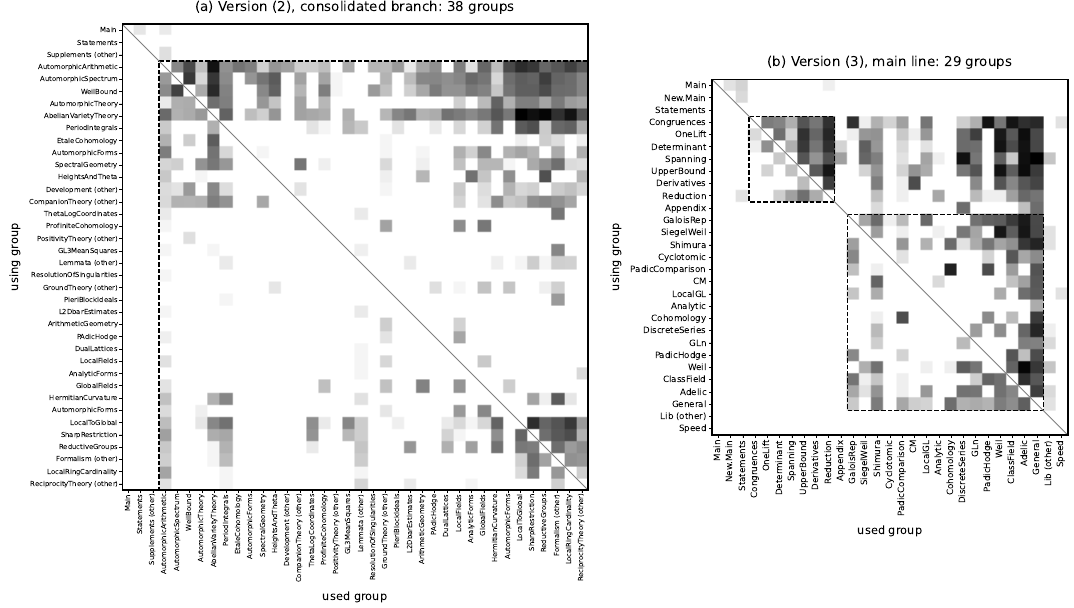}
\caption{Dependency structure matrices of namespace groups in (a) version~(2) and (b) version~(3). A row uses a column; the shade is the logarithm of the number of uses (pairs of a module and a name that it uses) between the two groups. Groups are ordered from \texttt{Main}, the module of the five theorems, downward by a heuristic that reduces uses below the diagonal (backward uses). Dashed squares enclose strongly connected sets of groups. In (b), the two squares are the proof-specific part (\texttt{New}, upper) and the library (\texttt{Lib}, lower); no counted use goes from the library to the proof-specific part. For version~(2), the 28 largest second-level namespaces and the statement files are shown, and the other namespaces are merged per top level.}
\label{fig:dsm}
\end{figure*}

\begin{figure}[!t]
\centering
\includegraphics[width=\columnwidth]{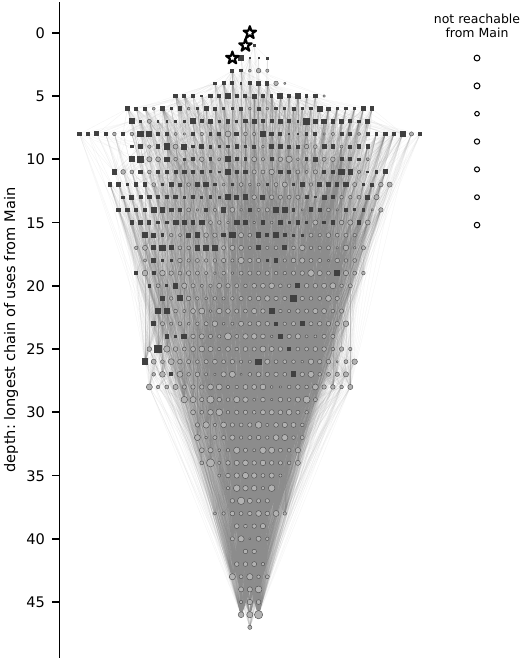}
\caption{The 802 modules of version~(3), placed by the length of the longest chain of uses from \texttt{Main} (top, depth~0). Dark squares: the 302 reachable modules under \texttt{New} other than \texttt{New.Main}; grey circles: the 490 reachable modules under \texttt{Lib}; stars: \texttt{Main}, \texttt{New.Main} and \texttt{Statements}; hollow circles at the right: the seven modules not reachable from \texttt{Main} (their vertical position has no meaning). The area of a mark grows with the number of real declarations in the module; a line joins two modules with at least one counted use, and the number of uses is not shown. Modules in the same row are ordered to reduce crossings.}
\label{fig:mod}
\end{figure}

To compare how the versions are organised, we aggregated the uses recorded in the compiled indexes (U rows: a module uses a name declared in another module) to modules and to namespace groups, a group being the first two components of the module name. Each used name was attributed to the module of its real declaration. Names declared in more than one module (320 in version~(3), all of them auxiliary lemmas such as \texttt{congr\_simp} and \texttt{eq\_1}, which Lean generates where they are needed; 8{,}244 in version~(2), of which 8{,}160 are such auxiliary lemmas and 84 are ordinary declarations that occur in two modules) were not counted. The uses counted in this way go from a module to modules that it imports, so at the level of modules they form no cycles; cycles appear only when modules are grouped. We call a set of groups strongly connected if each group in it reaches every other by a chain of uses.

Version~(2) has 47{,}777 modules with declarations, and the longest chain of uses has 147 steps. Its seven top-level namespaces form one strongly connected set, and of its 315 second-level namespaces, 86 form one strongly connected set. Version~(3) has 802 modules and a longest chain of 47 steps. Its two top-level namespaces are layered: modules under \texttt{New}, the proof-specific part, use names declared under \texttt{Lib} 30{,}086 times, and none of the counted uses goes from a module under \texttt{Lib} to a name declared under \texttt{New}. Within each layer the second-level groups are again strongly connected (all 7 groups under \texttt{New}, and 16 groups under \texttt{Lib}), so the library itself is not layered by namespace. Fig.~\ref{fig:dsm} shows the two structures as dependency structure matrices; for readability, version~(2) is shown with its 28 largest second-level namespaces and the statement files, the other namespaces being merged per top level (38 groups, of which 35 form one strongly connected set).

Fig.~\ref{fig:mod} shows the 802 modules of version~(3) themselves, each placed at the length of the longest chain of uses from \texttt{Main}. Seven modules are not reachable from \texttt{Main}: five under \texttt{Lib.ClassField} and two under \texttt{New.Congruences.Selmer}. Of the others, the 303 proof-specific modules (under \texttt{New}) lie at depths 1 to 27 (median 11, quartiles 7 and 14), and the 490 library modules (under \texttt{Lib}) at depths 3 to 47 (median 23, quartiles 17 and 29); the remaining two are \texttt{Main} and \texttt{Statements}. The two ranges overlap, but the proof-specific modules are concentrated in the upper part, and 162 library modules lie deeper than any proof-specific module.

The difference lies in the arrangement more than in the amount of backward dependence. With the groups of Fig.~\ref{fig:dsm} ordered by a simple heuristic (strongly connected sets in dependency order, and adjacent groups within them swapped while this reduces backward uses), 3.5\% of the uses between groups point backwards in version~(2) and 4.6\% in version~(3). These shares depend on the ordering and on the grouping, and are given only to show that both are small; the difference is that in version~(3) all backward uses lie inside one of the two layers. The groups are the authors' namespaces, so this comparison describes how the code is named and arranged, not a mathematical decomposition. Together with C1, it shows that version~(3) differs from version~(2) not only in its declarations but also in its organisation: a library with no counted use of the proof-specific part, and the proof built on top of it.

\section{Discussion, limitations, and use of generative AI}\label{sec:discussion}

\subsection{Two kinds of reduction}\label{sec:disc-two}

The three versions are connected by two reductions of a different kind.
From version~(1) to version~(2), volunteers consolidated the original release~\cite{alpoge_archive} on the branch \texttt{opt/collapse-and-consolidate-4.36} of \texttt{rohanarun/\allowbreak riemann-hypothesis}~\cite{repo_ra,pr120}.
The number of files fell from 175,187 to 47,885 and the number of lines from 17,374,215 to 14,583,569, but names were kept: 804,569 of the 1,057,984 named declarations in the source of version~(1) (76.05\%) have an exact counterpart in version~(2), almost all of them under the same fully qualified name, and only 6,858 (0.65\%) share no more than the last component of the name.
Within the limits of textual matching (Section~\ref{sec:disc-text}), what disappeared between the two versions was deleted rather than renamed: only 8 declarations of version~(1) match a declaration of version~(2) by statement under a different name.
From version~(2) to version~(3), the number of lines fell from 14,583,569 to 478,591 and the number of real declarations from 871,784 to 26,019, but almost nothing was kept at the level of declarations: only 47 declarations of version~(3) (0.18\%) have an exact counterpart in version~(2) (C1), and 9 of these are the five main theorems and the four definitions of the statement file, which the NOTICE of version~(3) says were taken over unchanged; of the remaining matches, most are short generic statements such as \texttt{: Type}, and only 15 matches with a statement of 20 or more characters remain.
That version~(3) is a rewriting is not our finding; its NOTICE describes it as ``a rewriting of Levent Alp\"oge's Lean proof''~\cite{repo_lc,repo_grhlean}.
What this paper adds is how far the rewriting goes at the level of declarations.

Two further observations support C1 without being separate claims.
First, the declarations of version~(2) that have no counterpart in version~(3) are used in the same way as version~(2) as a whole: 46.2\% of them are used from another file, 51.9\% are false zeros and 1.87\% are true zeros, against 47.9\%, 50.3\% and 1.84\% for all real declarations of version~(2) (C2).
Of the 5,507 declarations used from 100 or more modules, 17 have a counterpart in version~(3).
A reduction that removed unused declarations would leave a different profile; this one hardly selects on use (heavily used declarations are only slightly more likely to have a counterpart).
Second, the reduction is in the number of declarations, not in their length: for declarations with a statement line, the mean length of the declaration span (from the declaration line to the next line beginning in column~0) is 13.54 lines in version~(2) and 13.60 lines in version~(3) (C4).
The use of external constants per real declaration is also similar (29.5 and 28.4 uses of a pair of a module and an external constant), and 82.6\% of the external constants used by version~(3) were also used by version~(2).
The rewritten text is shorter in that it has fewer declarations, each about as long as before.

For Eduinformatics, the two reductions correspond to two ways of revising a textbook: editing an existing text by removing and merging parts, and writing a new text for the same goal.
Both produce a shorter text, and line counts alone do not distinguish them.
Tracing declarations across versions, as here, separates them, in the same sense in which our first paper examined the primary record of every declaration rather than aggregate counts~\cite{takamatsu2026unused}.

\subsection{The limits of pruning}\label{sec:disc-prune}

Whether version~(2) could have been reduced to a size close to that of version~(3) by deletion alone is bounded by the lines of version~(2) that are reachable from the main theorems (C3).
In version~(2), 826,654 of the 871,784 real declarations (94.8\%) are reachable, and they occupy 10,432,860 of its 10,737,061 declaration lines (97.2\%); version~(3), with 278,198 declaration lines (478,591 lines in all), has 2.67\% of this reachable part, about 1/37.5.
The reachable part is thus far larger than version~(3): deleting everything unreachable would remove only 2.83\% of the declaration lines of version~(2), and almost all of the reduction to version~(3) came from rewriting, not from deletion.
The lines of the reachable part are a lower bound on the size that pruning alone can reach, in the sense of deleting whole declarations on this compiled dependency graph without changing the remaining declarations or their proofs; it is not an estimate of the size that pruning would give. Shortening proofs, merging lemmas or inlining definitions are not pruning in this sense; they are forms of rewriting.
We do not claim that an unreachable declaration can be deleted: our first paper showed that a declaration with no recorded use can still be needed by the build~\cite{takamatsu2026unused}, and Section~\ref{sec:disc-invisible} gives the same observation for version~(3).
Conversely, the reachable part is not the smallest proof: a reachable declaration may be replaceable by a shorter one, which is what a rewriting does.
The bound therefore separates the two reductions only in one direction: the difference between the reachable lines of version~(2) and the lines of version~(3) is a reduction that deletion could not have produced.

Version~(3) itself is close to its own bound.
Of its real declarations in the dependency graph, 25,613 of 26,018 (98.44\%) are reachable from the five main theorems and 405 are not.
Apart from the three of the five main theorems that are true zeros and one true zero that does not appear in the dependency graph, every true zero of version~(3) is unreachable. Independent counts on \texttt{lsernau/grh-lean} point the same way, although they count differently: Nashida finds 550 of 24,092 declarations (counted by declaration names) unreachable from the five theorems, and 92\% of the declaration lines of its two main parts reachable~\cite{nashida2026star}; Muse reports a consumed dependency path of 8,850 declarations in 756 files~\cite{muse2026compiled}, a count of a different kind that is not comparable with our reachable real declarations.
For version~(1), three groups counted about 1.48 million declarations reachable from the main theorems (1,478,222~\cite{zhuk9581}, 1,478,243~\cite{yth4377} and 1,478,219~\cite{schiller3750}); these are counts of compiled constants and are not directly comparable with our real declarations, which fold auxiliary constants into the declaration they belong to.
A skeleton of the proof pruned by reachability has been extracted by others~\cite{repo_skeleton}; our contribution is to express the reachable part of version~(2) in lines and to compare it with version~(3).

\subsection{Textual matching is not mathematical identity}\label{sec:disc-text}

The correspondence between versions was found by matching names and normalized statements as text.
A textual match does not mean that two declarations have the same type: the same text can mean different things under different \texttt{open} and \texttt{variable} contexts.
Conversely, the same mathematics written under a different namespace, with different type names or with arguments in another order, does not match. This is a general property of syntactic comparison: in a study of proof dissimilarity in a proof kernel and in Lean/Mathlib, raw syntax could not tell a rewrite from a new argument, and equality of normal forms never held between differently written versions of the same argument~\cite{nikolaou2026proofs}.
Version~(3) places its declarations under new namespaces such as \texttt{New.*} and redefines types, so textual matching can only find declarations that were copied.
We therefore do not report how many lemmas of version~(2) survive \emph{as mathematics} in version~(3).
Category~E (declarations of version~(3) whose short name of at least 15 characters also occurs in version~(2) but whose statement differs; 385 declarations) is given only as a rough upper bound for rewritten counterparts; for example, \texttt{trace\_pow\_eq\_sum\_roots} and \texttt{theta\_eq\_zero\_iff\_xi\_dvd} have the same names in both versions but rewritten arguments and types.
We checked a random sample of 40 of the 385 category-E declarations by hand against the declarations of version~(2) with the same short name: only 2 (5\%; Wilson 95\% interval 1--17\%) state the same result in rewritten form, 20 are closely related but changed statements (generalised, specialised, or about an analogous object), and 18 (45\%; 31--60\%) are unrelated results that merely share a name, often a generic one such as \texttt{integrable\_mul\_conj}. Category~E therefore overstates rewritten counterparts considerably. The two same-statement cases differ from their counterparts only in binder brackets or in the names of bound variables, which our textual normalisation does not absorb; extrapolated to the 385 declarations (about 5--64), such cases would raise the 47 strict counterparts of C1 to at most about 111, still below 0.5\% of version~(3). In 99 of the 385 declarations no statement was extracted (mostly structure fields and constructors), so they fall into category~E by name alone.
Matching is also between sets, not one-to-one: for example, 30,270 declarations of version~(2) and 489 of version~(3) fall in category~C, where the statement is a generic type such as \texttt{: Prop} and the match is almost always accidental; these are not counted as correspondences.
Statements were cut at \texttt{:=}, at 700 characters and at 15 lines, and declarations without a statement line (structure fields, constructors, unnamed instances, declarations generated by macros) can enter only the matches by name.
For version~(1) we counted named declarations in the source, because version~(1) was not compiled; its counts are therefore not real declarations in the sense used for versions~(2) and~(3).

\subsection{Uses that the compiled index cannot see}\label{sec:disc-invisible}

Both the use profile (C2) and reachability (C3) are read from the compiled index, which records a use when a declaration occurs as a constant in the compiled form of another.
Some uses leave no constant.
A lemma proved by \texttt{rfl} and used in \texttt{simp [foo]} or \texttt{rw [foo]} can be discharged by definitional unfolding, so \texttt{foo} does not occur in the compiled term, and the index records no use of it~\cite{takamatsu2026unused}.
We measured this effect on version~(3).
Of its 326 true zeros (real declarations with no recorded use in any file), deletion was attempted, singly or in clusters, with a result for 114: 111 failed and three (\texttt{coe\_one}, \texttt{finAdele\_add\_apply} and \texttt{finAdele\_zero\_apply}, each deleted on its own) passed the full build.
In 69 of the 111 failures the deleted name itself was reported as an unknown identifier or constant, that is, the name was used in the source although it does not occur in the compiled terms.
The other 42 failed without naming the deleted declaration (for example, unsolved goals or \texttt{simp made no progress}); these are probably uses through attributes such as \texttt{@[simp]}, but side effects of the deletion cannot be fully excluded.
All 31 true zeros tried one at a time in a first, separate round failed.
Of the remaining true zeros, 146 could not be located by the deletion script and 66 were not tried.
Since the true zeros of version~(3) are, apart from the main theorems and one declaration missing from the dependency graph, all unreachable, these failures are direct examples of unreachable declarations that cannot simply be deleted.
It is also the reason why C3 is stated as a lower bound for pruning and not as a size that pruning would achieve.

\subsection{Pruning of version~(3) by many contributors at once}\label{sec:disc-concurrent}

Version~(3) is itself being pruned, by many contributors at the same time.
On 27 November 2026 at 12:20 UTC, against commit 7c4ccbd64d of the main line (805 Lean files), 1,045 pull requests were open on \texttt{leanprover-community/\allowbreak riemann-hypothesis}~\cite{repo_lc}.
Of these, 1,007 changed at least one Lean file present on the main line; 789 of the 805 files were changed by at least one open pull request, 661 by two or more and 174 by five or more.
Of six pull requests that one of the authors opened on 26--27 November to delete declarations with no recorded use, five changed the same declarations as earlier pull requests by other contributors and were closed, either as duplicates or because the same deletion had already been merged.
On the same commit, the compiled index gave 38 candidates for deletion (true zeros, excluding the statement files, instances, structures and declarations with attributes such as \texttt{@[simp]}); 11 of them were already deleted by an open pull request, and the names of all other 27 occur in the source outside their own declaration, which is the pattern that Section~\ref{sec:disc-invisible} associates with failed deletions.
In this sense, pruning of version~(3) guided by the compiled index was saturated at this snapshot: what the compiled index reports as unused is either already being removed or used in a way that the index does not record.
This agrees with the observation that version~(3) is close to its own bound (Section~\ref{sec:disc-prune}): further reduction of a proof of this kind would have to come from rewriting, not from deletion.
For Eduinformatics, the observation concerns the organisation of revision rather than its method: when many contributors apply the same check to the same text at the same time, the part of the revision that the check can find is used up quickly, and coordination between contributors, not the check itself, becomes the limit.
These counts describe a single moment of a repository that changes by the hour, and they include pull requests opened by one of the authors.

\subsection{What cannot be inferred about the rewriting}\label{sec:disc-how}

Our data describe the result of the rewriting, not how it was done.
The README of the repository on which version~(3) is based states that it was produced ``in collaboration with my colleagues at Anthropic, and Claude''~\cite{repo_grhlean}. The README of its readable edition adds that the rewriting follows the method of the original proof, that many steps are proved by other arguments, and that parts of the original, mostly background material, were ported and then rewritten; the README of the repository also states that no paper accompanies it (checked on 28 November 2026). The small number of exact counterparts (C1) is consistent with this description, but our counts cannot tell which declarations were ported and then rewritten. Nashida, comparing the shortened proof with the original by name, found that none of 19 representative names of \texttt{lsernau/grh-lean} occurs as a full name in the original~\cite{nashida2026star}.
We do not know how the work was divided between people and the AI system, how many attempts were made, or whether version~(2) or version~(1) was read during the rewriting; nothing in the counts above identifies any of these.
The observations that the new text uses the conventions of Mathlib (more \texttt{lemma} than \texttt{theorem}; 56\% of the real declarations of version~(3) are lemmas), that its \texttt{ATTRIBUTION.md} lists 126 declarations modelled on Mathlib, and that it depends on a library, TauCeti, that version~(2) does not use, are consistent with a rewriting that took Mathlib rather than version~(2) as its model, but they do not establish it.
In particular, we make no claim about why the rewritten proof could be written with fewer declarations.

\subsection{Limitations}\label{sec:limitations}

\paragraph{Snapshots.}
All results refer to three fixed commits: 40e56bab02 (the original release of 20 November 2026), e497a66 of \texttt{rohanarun/\allowbreak riemann-hypothesis}, and 3b68810 of \texttt{leanprover-community/\allowbreak riemann-hypothesis} (26 November 2026, 22:14 UTC).
Version~(3) is still changing, and other intermediate states existed that we do not study, for example a cleaned tree of 149,397 files and 15,168,600 lines on the community repository before it switched to the rewriting~\cite{sarihan2026newmath}.
At an earlier commit, c5451bd, the share of false zeros among the declarations with no cross-file use was 97.0\% and there were 336 true zeros, against 97.1\% and 326 at 3b68810; the size of later changes has not been measured.
The organisation of Section~\ref{sec:res-org} was recomputed in the same way at commit 7c4ccbd64d of the main line (27 November 2026): it has the same 802 modules, 26,171 real declarations, longest chain of 47 steps and the same two strongly connected sets of groups (16 and 7), with 30,088 counted uses from \texttt{New} to \texttt{Lib} and none from \texttt{Lib} to \texttt{New}.

\paragraph{Two ways of counting version~(3).}
The real declarations of version~(3) were counted in two ways: from the index in the same form as for version~(2) (26,019) and from the edges of its dependency graph (26,018).
The two differ by two automatically generated congruence lemmas (\texttt{*.hcongr\_N}); all other declarations and their classes of use agree.
The correspondence with version~(2) (C1) and reachability were computed on the dependency graph, so their denominator is 26,018.
The rule that folds auxiliary constants into real declarations is the one of our first paper~\cite{takamatsu2026unused}; it treats \texttt{hcongr\_N} lemmas as real declarations, of which version~(3) has 26 and version~(2) none, probably because of the different Lean versions (v4.36.0 and v4.37.0-rc1).

\paragraph{Different Lean and Mathlib versions.}
Versions~(2) and~(3) were compiled against different commits of Mathlib (fc4f59e and 31ce06e, 153 commits apart).
Of the 69,633 external names used by version~(2), 16 (0.023\%) do not exist in the environment of version~(3), so the comparison of external constants is not affected in practice; the comparison of declaration names inside the projects does not depend on Mathlib.
For version~(2) the index records uses of external constants only per module, not per declaration, so external dependence is compared per module for both versions.

\paragraph{Version~(1) was not compiled.}
Compiling version~(1), with 3.7 times as many files as version~(2), was not feasible within this study.
Its declarations were taken from the source by a parser that approximates Lean's naming, and statements were extracted for 1,042,271 of them.

\paragraph{The authors as contributors.}
One of the authors has opened pull requests on the repository of version~(2), as reported in our first paper~\cite{takamatsu2026unused}, and on that of version~(3).
This should be kept in mind when reading the paper.

\paragraph{Namespace groups.}
The comparison of organisation (Section~\ref{sec:res-org}) depends on the namespaces chosen by the authors of each version and on the grouping by the first two components of the module name. A different grouping would give different cycles; only the absence of counted uses from \texttt{Lib} to \texttt{New} in version~(3) does not depend on the grouping below the top level. The ordering of Fig.~\ref{fig:dsm} is heuristic, so the shares of backward uses depend on it.

\subsection{Data availability}
The inputs are public: the three repositories and commits named above identify them completely.

\subsection{Use of generative AI}
The scripts that extract declarations and statements, match them across versions, count uses and reachability, and attempt deletions, as well as the analyses, were made with Claude Code using the model Claude Opus 5.5 (Anthropic; November 2026), working under the direction of the authors.
The manuscript was also drafted with Claude Opus 5.5.
The numbers reported here were taken from the outputs of the scripts, not from text generated by the AI system; the main numbers were recomputed by a second, independently written script.
No generative AI system is an author, and the authors take full responsibility for the content of this paper.

\section*{Acknowledgment}
We thank Fumihiko Nakamura (Kitami Institute of Technology), Yuri Yatagawa (Institute of Science Tokyo), and Shun'ichi Honda (Chitose Institute of Science and Technology) for discussions and comments, and Hibiki Ito (University of Helsinki and Kyoto University) for support.

\bibliographystyle{IEEEtran}
\bibliography{main}

\begin{thebibliography}{10}
\providecommand{\url}[1]{#1}
\csname url@samestyle\endcsname
\providecommand{\newblock}{\relax}
\providecommand{\bibinfo}[2]{#2}
\providecommand{\BIBentrySTDinterwordspacing}{\spaceskip=0pt\relax}
\providecommand{\BIBentryALTinterwordstretchfactor}{4}
\providecommand{\BIBentryALTinterwordspacing}{\spaceskip=\fontdimen2\font plus
\BIBentryALTinterwordstretchfactor\fontdimen3\font minus
  \fontdimen4\font\relax}
\providecommand{\BIBforeignlanguage}[2]{{%
\expandafter\ifx\csname l@#1\endcsname\relax
\typeout{** WARNING: IEEEtran.bst: No hyphenation pattern has been}%
\typeout{** loaded for the language `#1'. Using the pattern for}%
\typeout{** the default language instead.}%
\else
\language=\csname l@#1\endcsname
\fi
#2}}
\providecommand{\BIBdecl}{\relax}
\BIBdecl

\bibitem{alpoge_archive}
L.~Alp{\"o}ge and {Claude}, ``Lean~4 archive \texttt{lean.tar.gz} (14 parts)
  and accompanying remark,'' \url{https://alpo.ge/}, Nov. 2026, released 20
  November 2026.

\bibitem{leanfro_blog}
{Lean FRO and Mathlib Initiative}, ``Lean verification of the {Riemann
  Hypothesis} (joint statement),'' Lean community blog,
  \url{https://leanprover-community.github.io/blog/posts/riemann-hypothesis/},
  2026, 21 November 2026.

\bibitem{repo_ra}
``rohanarun/riemann-hypothesis,'' GitHub repository,
  \url{https://github.com/rohanarun/riemann-hypothesis}, commit e497a66 on
  branch opt/collapse-and-consolidate-4.36. Accessed 26 November 2026.

\bibitem{pr120}
R.~Arun, ``Collapse 28,169 thin wrappers and consolidate 93,601 modules into
  20,258 bundles ($-101{,}512$ files, $-67.9\%$),'' Pull request \#120,
  leanprover-community/riemann-hypothesis,
  \url{https://github.com/leanprover-community/riemann-hypothesis/pull/120},
  2026, opened 22 November 2026 from branch
  rohanarun:opt/collapse-and-consolidate-4.36.

\bibitem{takamatsu2026unused}
K.~Takamatsu, K.~Murakami, S.~Matsumoto, T.~Akiyama, S.~Matsumoto, K.~Kunida,
  S.~Takahashi, T.~Matsuo, S.~Hirokawa, K.~Tsuruta, R.~Kozaki, S.~Nishida,
  K.~Takamatsu, F.~Lumban~Gaol, and M.~Mori, ````{N}ot used elsewhere'' is not
  ``unused'' in a {Lean} formalization of the {Riemann Hypothesis}: {R}everse
  lookup against the compiled index, based on {E}duinformatics,''
  arXiv:2611.37167 [cs.AI], 2026.

\bibitem{repo_lc}
``leanprover-community/riemann-hypothesis,'' GitHub repository,
  \url{https://github.com/leanprover-community/riemann-hypothesis}, branch
  main, commit 3b68810 (26 November 2026, 22:14 UTC).

\bibitem{repo_grhlean}
``lsernau/grh-lean,'' GitHub repository,
  \url{https://github.com/lsernau/grh-lean}, readable edition \texttt{grh/};
  version (3) is based on its commit 72fead1. NOTICE: ``a rewriting of Levent
  Alp{\"o}ge's Lean proof''; README: ``produced in collaboration with my
  colleagues at Anthropic, and Claude''. Accessed November 2026.

\bibitem{repo_rhshrink}
``rh-shrink/rh-shrink.github.io,'' GitHub repository,
  \url{https://github.com/rh-shrink/rh-shrink.github.io}, accessed November
  2026.

\bibitem{repo_novusedge}
``Novusedge/rh-proof-notes,'' GitHub repository,
  \url{https://github.com/NovusEdge/rh-proof-notes}, accessed November 2026.

\bibitem{repo_foxfire}
``Foxfire1st/rh-proof-skeleton,'' GitHub repository,
  \url{https://github.com/Foxfire1st/rh-proof-skeleton}, accessed November
  2026.

\bibitem{repo_engram}
``engram-agents/riemann-explained,'' GitHub repository,
  \url{https://github.com/engram-agents/riemann-explained}, accessed November
  2026.

\bibitem{zhuk9581}
I.~M. Zhuk, ``The {Riemann Hypothesis}: Mathematical exposition and kernel
  blueprint of the formal proof (edition 9.0), with a one-page blueprint in
  {Russian},'' 2026, 23 November 2026.

\bibitem{yth4377}
{yth (yuyoyth)}, ``Principled logical self-collision theorems, static quotient
  cross-section distortions, and source-code and dependency-graph verification
  of a seventeen-million-line dependent-type formalization of {GRH},'' 2026, 22
  November 2026.

\bibitem{schiller3750}
E.~Schiller, ``Does a trivially true predicate invalidate a kernel-checked
  proof? {A}n independent build and dependency audit of the {Alp{\"o}ge} {Lean}
  archive for the {Riemann Hypothesis},''
  \url{https://zenodo.org/records/24543750}, 2026, 25 November 2026.

\bibitem{takamatsu2026coverage}
K.~Takamatsu, K.~Murakami, S.~Matsumoto, T.~Akiyama, S.~Matsumoto, K.~Kunida,
  S.~Takahashi, T.~Matsuo, K.~Tsuruta, R.~Kozaki, S.~Nishida, K.~Takamatsu,
  F.~Lumban~Gaol, and M.~Mori, ``How does coverage of forward links affect the
  reliability of reverse lookup, and how far can it be corrected?
  {E}dge-omission experiments and correction of false ``never cited'' counts on
  citation graphs of mathematics textbooks based on {E}duinformatics,'' Zenodo
  preprint, doi: 10.5281/zenodo.24556911, 2026.

\bibitem{edu2018}
K.~Takamatsu, K.~Murakami, T.~Kirimura, K.~Bannaka, I.~Noda, R.~W. Lim
  \emph{et~al.}, ``Eduinformatics: A new education field promotion,''
  \emph{Bulletin of Kobe Tokiwa University}, vol.~11, pp. 27--43, 2018.

\bibitem{takamatsu2026review}
K.~Takamatsu, ``The first week of research on the {Lean} proof of the {Riemann
  Hypothesis}: A review of preprints, 20--26 {N}ovember 2026,'' Zenodo
  preprint, doi: 10.5281/zenodo.24557728, 2026.

\bibitem{roy2009clones}
C.~K. Roy, J.~R. Cordy, and R.~Koschke, ``Comparison and evaluation of code
  clone detection techniques and tools: A qualitative approach,'' \emph{Science
  of Computer Programming}, vol.~74, no.~7, pp. 470--495, 2009.

\bibitem{kim2005genealogies}
M.~Kim, V.~Sazawal, D.~Notkin, and G.~Murphy, ``An empirical study of code
  clone genealogies,'' in \emph{Proceedings of the 10th European Software
  Engineering Conference held jointly with the 13th ACM SIGSOFT International
  Symposium on Foundations of Software Engineering (ESEC/FSE)}, 2005, pp.
  187--196.

\bibitem{gode2013evolution}
N.~G{\"o}de and R.~Koschke, ``Studying clone evolution using incremental clone
  detection,'' \emph{Journal of Software: Evolution and Process}, vol.~25,
  no.~2, pp. 165--192, 2013.

\bibitem{weiser1984slicing}
M.~Weiser, ``Program slicing,'' \emph{IEEE Transactions on Software
  Engineering}, vol. SE-10, no.~4, pp. 352--357, 1984.

\bibitem{ringer2019qed}
T.~Ringer, K.~Palmskog, I.~Sergey, M.~Gligoric, and Z.~Tatlock, ``{QED} at
  large: A survey of engineering of formally verified software,''
  \emph{Foundations and Trends in Programming Languages}, vol.~5, no. 2--3, pp.
  102--281, 2019.

\bibitem{mathlib2020}
{The mathlib Community}, ``The {Lean} mathematical library,'' in
  \emph{Proceedings of the 9th ACM SIGPLAN International Conference on
  Certified Programs and Proofs (CPP)}, 2020, pp. 367--381.

\bibitem{vandoorn2020maintaining}
F.~van Doorn, G.~Ebner, and R.~Y. Lewis, ``Maintaining a library of formal
  mathematics,'' in \emph{Intelligent Computer Mathematics (CICM 2020)}, ser.
  Lecture Notes in Computer Science, vol. 12236, 2020, pp. 251--267.

\bibitem{nashida2026star}
H.~Nashida, ``What does the main inequality ($\star$) assert? {F}indings from
  examining the {S}ernau shortened proof of the {R}iemann hypothesis through
  its {Lean} code,'' Zenodo preprint, 2026, version r8, 27 November 2026.

\bibitem{muse2026compiled}
Muse, ``Compiled, not proved: A complete extraction and grading of the grh-lean
  certificate for the {R}iemann hypothesis,'' Zenodo preprint, 2026.

\bibitem{repo_skeleton}
``Foxfire1st/rh-proof-skeleton,'' GitHub repository,
  \url{https://github.com/Foxfire1st/rh-proof-skeleton}, accessed 27 November
  2026.

\bibitem{nikolaou2026proofs}
N.~Nikolaou, ``How far apart are two proofs? {M}easuring proof dissimilarity in
  a proof kernel and in {Lean}/{Mathlib},'' arXiv:2611.31811, 2026.

\bibitem{sarihan2026newmath}
T.~Sarihan, ``newmath-rgh: Reading guide to the new mathematics in {Alp\"oge}'s
  {Lean}-verified {R}iemann hypothesis proof,'' Zenodo record, 2026, version
  1.3.0.

\end{thebibliography}
\end{document}